**Title**:

Machine learning predicts long-term mortality after acute myocardial infarction using systolic time intervals and routinely collected clinical data

**Short title**:

Machine learning predicts long-term mortality after acute myocardial infarction

**Authors Names and affiliations**:

Bijan Roodini[1], BSc, Boshra Khajehpiri[1], MSc, Hamid Abrishami Moghaddam[2], PhD, and Mohamad Forouzanfar[1,3], PhD

[1]Department of Systems Engineering, École de technologie supérieure, Université du Québec, Montréal, QC, Canada

[2]Machine Vision and Medical Image Processing (MVMIP) Laboratory, Faculty of Electrical and Computer Engineering, K. N. Toosi University of Technology, Tehran, Iran

[3]Centre de recherche de l'Institut universitaire de gériatrie de Montréal (CRIUGM), Montréal, Canada

**Corresponding author**: Mohamad Forouzanfar, École de technologie supérieure, 100 Notre-Dame Street West, Montréal, QC H3C 1K3, Canada

Phone: +1-514-396-8720. Email: mohamad.forouzanfar@etsmtl.ca

# Abstract

**Background:** Precise estimation of cardiac patients' current and future comorbidities is an important factor in prioritizing continuous physiological monitoring and new therapies. Machine learning (ML) models have shown satisfactory performance in short-term mortality prediction of patients with heart disease, while their utility in long-term predictions is limited. This study aims to investigate the performance of tree-based ML models on long-term mortality prediction and the effect of two recently introduced biomarkers on long-term mortality.

**Methods:** This study utilized publicly available data from the Collaboration Center of Health Information Application (CCHIA) at the Ministry of Health and Welfare, Taiwan, China[1]. The data pertained to patients admitted to the cardiac care unit for acute myocardial infarction (AMI) between November 2003 and September 2004. Included in the dataset were patients over 20 years old diagnosed with type 1 AMI. Exclusion criteria encompassed patients with missing data for brachial pre-ejection period (bPEP) and brachial ejection time (bET), as well as those with atrial fibrillation or extremity amputations, resulting in a final cohort of 139 AMI patients. We collected and analyzed mortality data up to December 2018. All patients had provided informed consent, in accordance with the Declaration of Helsinki. Medical records were used to gather demographic and clinical data, including age, gender, body mass index (BMI), percutaneous coronary intervention (PCI) status, and comorbidities such as hypertension, dyslipidemia, ST-segment elevation myocardial infarction (STEMI), and non-ST segment elevation myocardial infarction (NSTEMI). Using medical and demographic records as well as two recently introduced biomarkers, bPEP and bET, collected from 139 patients with acute myocardial infarction, we investigated the performance of advanced ensemble tree-based ML algorithms (random forest, AdaBoost, and XGBoost) to predict all-cause mortality within 14 years. A nested cross-validation was performed to evaluate and compare the performance of our developed models precisely with that of the conventional logistic regression (LR) as the baseline method.

**Results:** The developed ML models achieved significantly better performance compared to the baseline LR (C-Statistic, 0.80 for random forest, 0.79 for AdaBoost, and 0.78 for XGBoost, vs 0.77 for LR) ($P_{RF}<0.001$, $P_{AdaBoost}<0.001$, $P_{XGBoost}<0.05$). Adding bPEP and bET to our feature set significantly improved the algorithms' performance, leading to an absolute increase in C-Statistic of up to 0.03 (C-Statistic, 0.83 for random forest, 0.82 for AdaBoost, and 0.80 for XGBoost, vs 0.74 for LR) ($P_{RF}<0.001$, $P_{AdaBoost}<0.001$, $P_{XGBoost}<0.05$).

**Conclusion:** The study indicates that incorporating new biomarkers into advanced machine learning models significantly improves long-term mortality prediction in cardiac patients. This advancement may enable better treatment prioritization for high-risk individuals[2].

---

# Introduction

Cardiovascular diseases are the leading cause of death globally, with an estimated 17.9 million deaths per year [1]. Approximately every 40 seconds, someone in the United States experiences a myocardial infarction (MI) [2]. In Canada, AMI patients are four times more likely to die of all causes in a given year compared to individuals without AMI [3]. In Europe, about half of the major coronary events occur in those with a previous hospital discharge of AMI, and 1 out of every 5 AMI patients suffers a second cardiovascular event in the first year [4]. Given the high risk of cardiovascular events following an MI, MI patients should be carefully monitored and managed with effective prevention programs.

Implantable sensors and smart wearables have provided in-home solutions for continuous monitoring of patients diagnosed with AMI [5, 6]. Given the fact that continuous monitoring of patients with AMI is expensive, requires a lot of resources, and is not available for every patient, it is wise to focus on patients with more serious health conditions and at higher risk of mortality. A precise estimation of the patient's current and future health situation is important to prioritize them for continuous physiological monitoring and to provide them with specific therapies at the earliest possible time [7].

Risk assessment in clinics, particularly for heart disease patients, often relies on methods like the Global Registry of Acute Coronary Events (GRACE) [8] score for acute coronary syndrome (ACS) patients, including AMI patients, and the Thrombolysis in Myocardial Infarction (TIMI) [9] score for MI patients. These scores, based on clinical information and factors like age and blood pressure, are validated for predicting short-term mortality but may not fully address long-term mortality risks in AMI patients.

Recently, machine learning (ML) has been employed as a data-driven approach to design accurate classifiers for the risk estimation of patients with cardiovascular diseases [10-19]. ML algorithms are able to reveal complex hidden patterns in data that cannot be extracted with traditional tools. Artificial neural networks (ANN) and tree-based methods have successfully been used to diagnose heart diseases [20-25] and predict CV outcomes [10-19]. Hernesniemi et al. [13] utilized XGBoost and logistic regression (LR) to predict the 6-month mortality of patients with ACS. They found that ML models outperformed the GRACE risk score system, with XGBoost achieving the best accuracy. Hadanny et al. [12] showed that the random forest (RF) algorithm performs drastically better than the GRACE score in post-ST-segment elevation MI (post-STEMI) 30-day mortality prediction. Cho et al. [11] reviewed several articles to compare the ability of ML and conventional statistical models to predict readmission and mortality in patients with AMI. Out of 19 articles, 13 demonstrated a higher performance for ML models compared to traditional techniques. Lee et al. [16] showed the outperformance of ML models over conventional methods, including GRACE and TIMI scores, in both short- and long-term mortality prediction of patients with a history of non-ST-segment elevation MI (NSTEMI). Li et al. [17] developed LR, ANN, and tree-based ML models to predict 1-year post-discharge mortality status of about 220,000 patients diagnosed with AMI. The constructed models reached average accuracies of 0.8-0.85, higher than the previously reported in-hospital mortality prediction models. Mansoor et al. [18] studied 10,000 women with STEMI to predict their in-hospital mortality using LR and RF. Their results revealed a comparable performance between the models and confirmed RF as a practical ML tool in clinical settings. Vomlel et al. [19] aimed to make a 30-day mortality prediction of 603 STEMI patients using different classifiers, including LR, Bayesian methods, ANN, and tree-based algorithms. Comparing their prediction power, LR and simple Bayesian methods were shown to be the most promising models. Other studies [10, 15] with different cohort sizes (5,000 and 22,000 patients) have tried to predict 1-year mortality after the diagnosis of AMI and have reported similar risk factors such as age and sex as the most important features.

Most existing studies have focused on short-time mortality prediction and building computational models without recognizing the most important risk factors. Prediction of long-term mortality is essential to prioritize patients who are at higher risk as earlier receivers of medications, preventive interventions, and healthcare plans. In this study, we investigate the application of state-of-the-art tree-based ML models using a set of easy-to-access clinical measures to predict long-term (14-year) mortality in 139 patients (52 survivors and 87 deaths) with AMI. In addition to routinely collected clinical measures, we consider recently introduced non-invasively measured systolic time intervals, the brachial pre-ejection period (bPEP), and brachial ejection time (bET) [26-28] as novel biomarkers for early prediction of all-cause mortality. The bPEP/bET ratio has been proven to be a useful variable for the risk assessment of AMI patients [29], and in this study, we aim to investigate its effectiveness on the performance of ML models. We study the correlation strength between different risk factors and the models' predictive power to find the most important mortality biomarkers in individuals suffering from AMI.

## Methods

*Study Population*

This analysis was based on publicly available data from the Collaboration Center of Health Information Application (CCHIA), Ministry of Health and Welfare, Taiwan, China [29]. The dataset comprised patients admitted to the cardiac care unit due to acute myocardial infarction (AMI) between November 2003 and September 2004. Inclusion criteria were patients older than 20 years diagnosed with type 1 AMI. Exclusion criteria included absence of data on brachial pre-ejection period (bPEP), brachial ejection time (bET), presence of atrial fibrillation, or amputation of extremities. This resulted in a dataset of 139 AMI patients (36 women) aged between 24-91 years. The study tracked and analyzed the mortality of these patients until December 2018, during which 87 patients died. In compliance with the Declaration of Helsinki, all patients provided informed consent for their data to be used in the study. Medical records were employed to gather demographic and medical data, including age, gender, body mass index (BMI), percutaneous coronary intervention (PCI) status, and comorbidities such as hypertension, dyslipidemia, ST-segment elevation myocardial infarction (STEMI), and non-ST segment elevation myocardial infarction (NSTEMI). The data from CCHIA were initially collected to assess the potential of the bPEP to bET ratio as a novel biomarker for predicting long-term cardiovascular and overall mortality in AMI patients [29]. For the purpose of our study, we dichotomized the cohort based on mortality status at the end of the study into two groups: those who survived and those who died within the 14-year period post-AMI diagnosis.

*Predictor Features*

The dataset contains 11 predictor variables, including clinical and demographic measurements acquired from each participant at baseline, containing five numeric and six binary features. The numeric variables consist of bPEP, bET, body mass index (BMI), ankle-brachial index (ABI), and age. The binary features include received percutaneous coronary intervention (PCI) or not, sex, and medical status of the patients on comorbidities such as dyslipidemia, diabetes, hypertension, and STEMI. ABI, bPEP, and bET were obtained using an ankle-brachial index ABI-form device recorded within 24 hours of each participant's admission [29]. Table 1 gives summary statistics of the data measurements at baseline comparing the mortal group to those who survived (non-mortals).

*Machine Learning*

We selected three advanced tree-based ML algorithms to perform our prediction task by finding an optimal decision boundary between the two groups of survivors and mortals. After building our models, they were tested to predict the mortality status of new individuals in the following 14 years.

Tree-based ML methods have proved their strength in modeling complex nonlinear input-output patterns in a nonparametric fashion, especially for structured data [30]. These algorithms achieve the best performance when used in an ensemble learning framework [31]. Ensemble-based machine learning algorithms work by running multiple base learners and aggregating the decisions made by each to achieve a more reliable outcome. These methods can usually handle noisy data and outliers well with little effect on the overall performance. They are also robust against overfitting and have lower variance than each of their weak learners [32]. Bagging and boosting methods form the two main categories of ensemble learning approaches that are particularly favored over more complex deep learning models in learning tasks where limited data are available.

*Random Forest*

Random forest (RF) is an ensemble of decision trees that makes predictions by averaging the decisions obtained from all its base learners [33]. RF benefits from two sources of randomization: first, the adoption of a random bootstrap of the data for each tree building, and second, restricting the candidate features to a random subset at each node splitting level of the trees. Averaging the results of multiple models along with the randomization helps the forest to maintain a low generalization error while growing deeper trees.

*Boosting Machines*

In the realm of advanced boosting techniques, Adaptive Boosting (AdaBoost) [34] and eXtreme Gradient Boosting (XGBoost) [35] stand out as notable extensions of traditional boosting methods. AdaBoost is characterized by its sequential ensemble model structure and minimal hyperparameter configuration, which lends it robustness against overfitting, particularly in scenarios involving low-noise and smaller datasets. On the other hand, XGBoost represents an efficient iteration of the gradient tree boosting methodology. Its primary strengths lie in its rapid processing speed, surpassing other similar algorithms, and its regularization feature, which significantly reduces variance and enhances model performance.

*Data Analysis*

The dataset we used did not contain any missing values. One-hot encoding was applied to convert the categorical features into binary variables [36]. To enhance the numerical stability of the models, standard scaling (z-score normalization) was applied by subtracting the mean from the feature values and then dividing them by the standard deviation [37].

Model selection, hyperparameter optimization, and performance evaluation were performed using nested cross-validation to verify the generalization ability of the ML models. This data division technique helps to estimate an unbiased generalization performance of a model and is specifically useful when dealing with smaller data sizes. The cross-validation method was implemented as follows: A 10-fold outer loop was first formed by dividing the whole data randomly into ten equal parts. For each fold, 90% of the samples entered a 5-fold inner loop for model selection and hyperparameter tuning, while the remaining 10% was set aside for final testing. This procedure was repeated ten times to reduce the inherent randomness of ML models, and the average results were reported. It is important to note that all the reported results are on the test set, which was not seen during training and validation.

Table 2 lists the hyperparameters, their search ranges, and their optimal values for each ML model. They include the general learning hyperparameters of the models (e.g., learning rate, input sampling method, and node splitting criteria) as well as those that determine model complexity and control underfitting/overfitting (e.g., number of estimators, maximum depth of the trees, pruning thresholds, and feature sampling rate at each node).

*Statistical Analysis*

Our designed ML models were evaluated based on the receiving operating characteristics (ROC) metric. The models were compared with each other using the area under ROC curve (AUC) index, also known as C-statistic, accuracy, sensitivity (true positive rate, recall), specificity, and precision.

Paired t-tests were performed on the cross-validation results to investigate whether the performances of different ML methods are superior to those of the baseline LR. A one-way repeated measures ANOVA was performed to compare the performances of the three ML models, followed by a Tukey post-hoc multiple comparisons.

The Gini feature importance method [38] was utilized to rank the most important predictive features. To investigate the effect of bPEP and bET on the prediction performance, the same experiments were repeated by excluding the two parameters from the analysis.

# Results

*Cohort Characteristics*

The study longitudinally followed 139 patients initially diagnosed with Acute Myocardial Infarction (AMI). Of these, 87 participants succumbed, while 52 survived until the final follow-up.

*Model Performance*

As depicted in Figure 1, the average ROC curves demonstrate the performance of the developed models. Table 3 details their predictive capabilities. Notably, the RF model exhibited the highest AUC of 0.83. AdaBoost excelled in accuracy (82%), sensitivity (90%), specificity (69%), and precision (79%), surpassing other models in these metrics. Compared to the baseline LR, all ML models showed significant enhancements across all classification metrics.

*Variable Importance*

Figure 2 presents the relative importance of the input variables in RF, AdaBoost, and XGBoost models. Age, BM, ABI, bET, and bPEP were identified as the most influential factors across all models. Conversely, variables such as diabetes, dyslipidemia, PCI, STEMI, and sex demonstrated the least predictive power.

*Effect of bPEP and bET*

Our analysis focused on assessing the impact of including brachial pre-ejection period (bPEP) and brachial ejection time (bET) in the predictive models. This involved retraining and testing the algorithms with and without these parameters, as detailed in Table 4 and Supplementary Tables 1 and 2.

In Experiment I, which included all features, the Random Forest (RF) model exhibited superior performance with an AUC of 0.83, an accuracy of 77%, sensitivity of 88%, specificity of 60%, and precision of 79% (all with p<0.001). AdaBoost also performed well with an AUC of 0.82, accuracy of 78%, and precision of 82%, while XGBoost had an AUC of 0.80. The training results for Experiment I, as shown in Supplementary Table 1, mirrored these findings, with RF and AdaBoost demonstrating high AUCs of $0.83 \pm 0.11$ and $0.82 \pm 0.10$, respectively, and XGBoost showing a slightly lower AUC of $0.81 \pm 0.12$.

In contrast, Experiment II, where bPEP and bET were excluded, showed a noticeable decline in the performance metrics for all models. The RF model's AUC decreased to 0.80, with corresponding drops in accuracy, sensitivity, specificity, and precision. AdaBoost and XGBoost also showed reduced performance, as reflected in their AUCs of 0.79 and 0.78, respectively. The training results for Experiment II, detailed in Supplementary Table 2, consistently revealed lower performance metrics across all models compared to Experiment I.

The inclusion of bPEP and bET significantly improved the RF model's performance across all classification metrics (p<0.001-0.01). AdaBoost showed marked improvements in AUC, accuracy, and sensitivity (p<0.001), and XGBoost demonstrated a significant increase in AUC (p<0.05) when these parameters were included. These results underscore the importance of bPEP and bET as predictive features in the models, contributing substantially to their accuracy and reliability in predicting mortality in AMI patients.

## Discussion

In this study, tree-based ML algorithms were developed for the mortality prediction of patients diagnosed with AMI over a long period of 14 years. Our models received clinical and demographic information as well as two non-invasively measured systolic time intervals, the bPEP and bET. Among different ML algorithms, ensemble tree-based ML algorithms were selected as they perform drastically better than other algorithms in most applications with limited available data

[31]. Bagging (RF) and two boosting (AdaBoost and XGBoost) ensemble algorithms were considered in this study.

It was observed that all the ML models significantly outperform the baseline LR algorithm. The best results were achieved using RF and AdaBoost, where RF achieved the highest C-statistic (0.83) and sensitivity (88%), and AdaBoost achieved the highest accuracy (47%), specificity (67%), and precision (82%). The AdaBoost classifier correctly predicted 78 mortal patients out of 87 mortal cases, which means that with a spanning of 70% of the study population, approximately 90% of mortal cases were distinguished. RF closely followed AdaBoost by correctly predicting 77 mortal patients out of the 87 mortal cases. XGBoost slightly underperformed compared to RF and AdaBoost.

Additionally, it is crucial to highlight the consistency in model performance across training and testing sets, as indicated in Supplementary Tables 1 and 2 for Experiments 1 and 2. The minor differences observed between these sets underscore the reliability and generalizability of the models, suggesting their robustness in practical applications.

A one-way ANOVA test was performed to find differences in the ML classifier's performances. A meaningful difference in sensitivity and specificity between ML models was observed ($p<0.05$). Tukey multiple comparison test was performed to find the superior models. It was observed that RF performs significantly better than XGBoost in terms of sensitivity, while AdaBoost and XGBoost perform significantly better than RF in terms of specificity. While AdaBoost achieved higher mean sensitivity compared to RF (88% vs 85%), it attained a higher standard deviation of sensitivity (13% vs 10%). A higher sensitivity means a lower false negative rate, which is extremely important in the prediction of mortality in AMI patients. Given that there was no significant difference between RF and AdaBoost models in terms of sensitivity, both algorithms can be considered equally suitable for our application. In terms of model complexity, AdaBoost is superior to RF as it only requires 20 estimators compared to 250 estimators for RF.

One of the objectives of this study was to examine the usefulness of the recently introduced features, bPEP and bET, in the prediction of long-term mortality of patients with AMI using ML models. PEP is the time it takes from the electrical depolarization of the left ventricle to the beginning of ventricular ejection, i.e., when the aortic valve opens. It is an index of myocardial contractility and beta-adrenergic sympathetic control of the heart. ET is the time it takes from the opening of the aortic valve to its closure and is conventionally used to evaluate the ventricle function and contractibility. Heart impairments usually prolong the PEP and shorten the ET [39, 40].

The higher values of bPEP/bET are shown to have a high correlation with cardiovascular mortality[29]. bPEP and bET can be easily calculated from the morphology of non-invasively measured pulse waveform and ECG and, therefore, can be easily integrated into any predictive model with minimum cost. According to Table 4, adding the bPEP and bET to our input feature set leads to the highest prediction performance, which implies their importance as new biomarkers of mortality in patients with AMI. It was observed that by considering bPEP and bET, the RF model's AUC, accuracy, sensitivity, specificity, and precision are improved by 3%, 4%, 2%, 8%, and 4%, respectively. For the AdaBoost model, AUC and precision were improved by 3% and 1%, respectively, while other classification metrics remained unchanged. It can, therefore, be concluded that RF can better extract the additional predictive information from the new features thanks to its more advanced architecture with a greater number of estimators.

Analyzing the importance of features in an ML model is another way of finding the most useful features for mortality prediction. Age, ABI, BMI, bET, bPEP, and hypertension were among the most important factors for mortality prediction. Most of these factors also had significant differences between mortal and non-mortal cases (Table 1). Our findings are in accordance to previous studies where age was an important factor in predicting the overall mortality of patients after AMI and heart failure [15, 41].

This study was limited to the application of tree-based algorithms for the prediction of long-term mortality of patients with AMI. ANN models, especially those with deep architectures, have recently dominated the computational biology field. ANN-based algorithms can perform better than other ML models when a large volume of training data is available. This study was limited to the analysis of data collected from 139 individuals, and therefore, tree-based algorithms were selected to avoid over-fitting and achieve a more generalizable assessment. Future work should be directed toward the collection of a larger and more detailed dataset where advanced deep learning models can be applied to predict a more precise time of mortality in the future. Another limitation of this study was the unbalanced sex distribution (36 women out of 139) while being a woman diagnosed with AMI is an important risk factor [42]. By collecting a balanced dataset of all genders, the effect of sex on the mortality of patients diagnosed with AMI can further be studied.

In clinical practice, evidence-based risk assessment methods are integral in distinguishing heart disease patients requiring intensive care unit level attention. Prominent among these is the GRACE score, a widely used tool for stratifying risk in patients with ACS [8]. Given that AMI is a subset of ACS, the GRACE score becomes pertinent for assessing the mortality/MI risk post-AMI. This score is derived from various clinical parameters, including age, heart rate, and systolic blood pressure. Its efficacy in predicting short-term (6-month to 1-year) all-cause mortality for AMI patients is well-established [43-47]. Another method, the TIM) score, is employed to estimate short-term mortality in MI patients, incorporating factors like age, coronary artery disease (CAD) risk factors, and clinical history [9]. Both GRACE and TIMI scores are thus pivotal in assessing short-term mortality risks. While our study's tree-based algorithms offer insightful data, they are not directly comparable to traditional GRACE and TIMI scores due to the lack of certain factors in our dataset necessary for these models. Nevertheless, future studies could explore a comparative analysis between these advanced tree-based algorithms and the traditional models like GRACE and TIMI.

It's important to note that the nested cross-validation method used in this study, although beneficial in reducing model overfitting, introduces minor dependencies between training and test sets. This interdependency slightly contravenes the paired t-test assumption of independence between folds, presenting a methodological limitation [48, 49]. Despite this, the impact on our results is expected to be minimal, owing to the rigorous nature of nested cross-validation [50].

## Conclusions

This study investigated the all-cause long-term mortality prediction power of different ensemble tree-based ML algorithms using routinely collected clinical data as well as two new biomarkers, bPEP and bET. It was found that ML models can accurately predict mortality with a C-statistic of as high as 0.83, statistically superior to LR as the baseline model. We also demonstrated that adding bPEP and bET to our input feature set improves the prediction results in most of the evaluation metrics. Predicting the mortality status of patients diagnosed with AMI over a long period of 14 years after AMI was another critical feature of this study. Our fast and easy-to-use ML models can assist medical staff in prioritizing patients with AMI for intense monitoring and preventing severe outcomes. Future research should be directed toward enhancing the prediction performance by collecting a larger balanced dataset and using more advanced ML algorithms such as ANNs.

## Conflicts of interest statement

The authors have no conflicts of interest to declare.

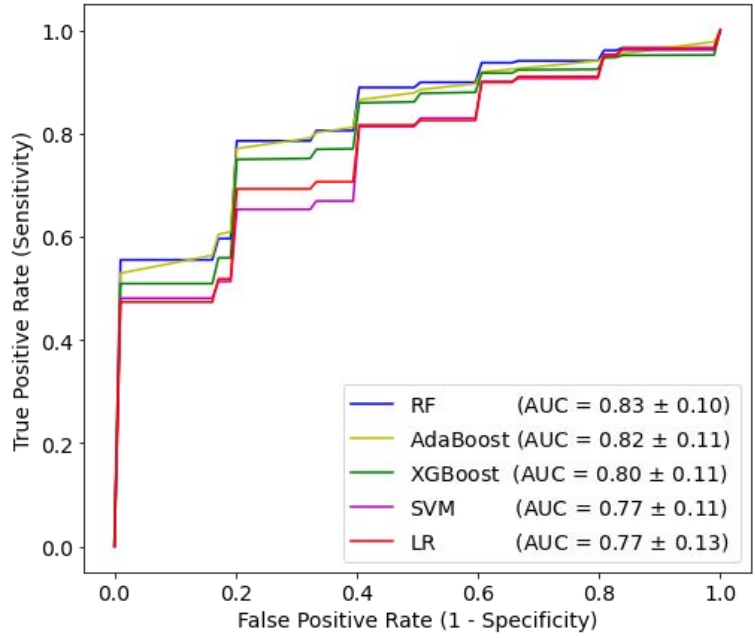

Figure 1. ROC curves for different AMI mortality prediction models.

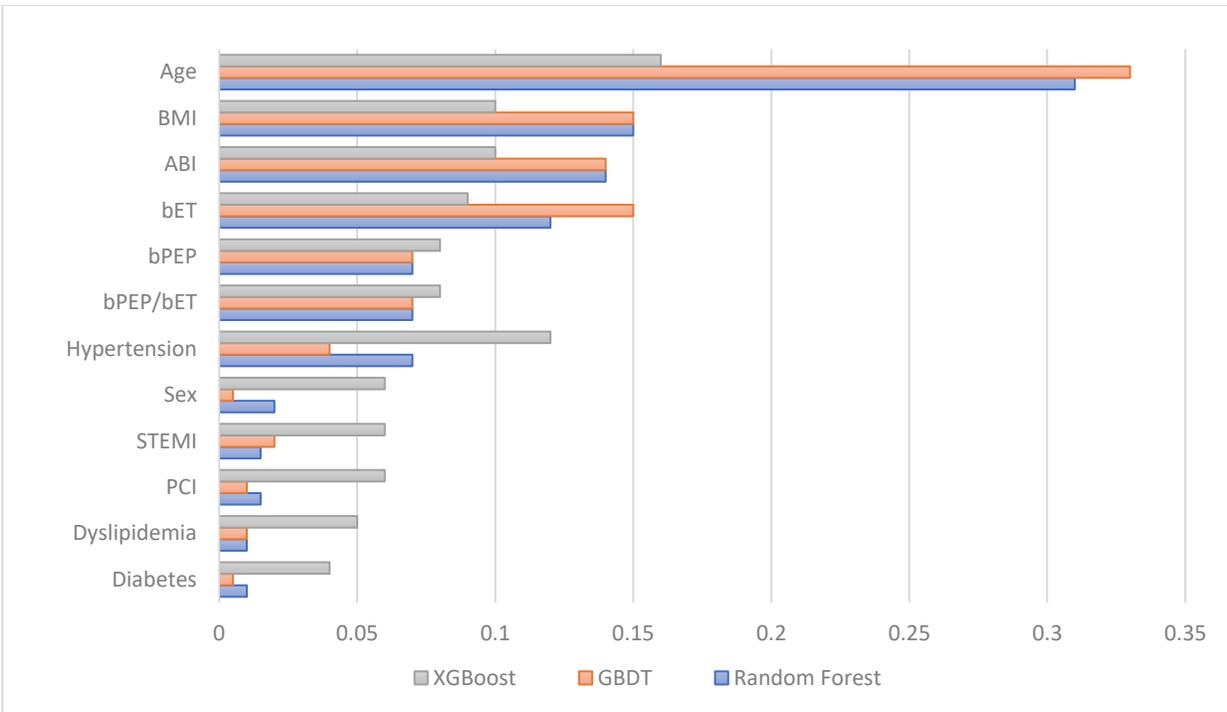

Figure 2. Variable importance for AMI mortality prediction.

Table 1. Statistics of Input Data

| Variables | Death Yes (N=87) | Death No (N=52) | P-Value |
|---|---|---|---|
| Age, (year) | 69 (12) | 55.3 (11.5) | <0.001* |
| Sex (male) | 60 (69%) | 43 (83%) | 0.112 |
| Diabetes mellitus | 21 (24 %) | 13 (25%) | 0.929 |
| Hypertension | 45 (27%) | 15 (29 %) | 0.014* |
| Dyslipidemia | 30 (34%) | 14 (27%) | 0.460 |
| PCI | 33 (38%) | 17 (33%) | 0.660 |
| STEMI | 18 (21%) | 12 (23%) | 0.906 |
| bPEP | 96.6 (21.0) | 94.8 (18.8) | 0.615 |
| bET | 250 (34.9) | 262.4 (24.1) | 0.025* |
| ABI | 0.94 (0.2) | 1.04 (0.1) | 0.001* |
| BMI | 23.6 (3.9) | 25.7 (3.1) | 0.002* |

Values are presented as mean (standard deviation) for continuous variables and n (%) for categorical variables
PCI: percutaneous coronary intervention; STEMI: ST-segment MI; bPEP: brachial pre-ejection period; bET: brachial ejection time; ABI: ankle brachial index; BMI: body mass index.
T-test and Chi-square test are performed on numerical and categorical variables, respectively, to examine the dependency between each variable in mortal and non-mortal cases.
*: significant difference (p<0.05) between the mortal and non-mortal groups.

Table 2. Hyperparameter tuning for ML algorithms.

| Model | Hyperparameters | Search interval | Optimum value |
| --- | --- | --- | --- |
| RF | Number of estimators | [20, 50, 150, 200, 250, 300, 350, 400, 500, 1000] | 250 |
|  | Max number of features | ['auto', 'sqrt'] | 'sqrt' |
|  | Max depth | [1, 3, 5, 6, 7, 12, 14, 16, 18] | 3 |
|  | Min samples split | [2, 4, 6, 8, 10, 12, 14] | 10 |
|  | Min samples leaf | [2, 3, 4, 5, 6, 7, 9, 12] | 6 |
|  | Bootstrap | [True, False] | False |
|  | Criterion | ['entropy', 'gini'] | 'entropy' |
| AdaBoost | Number of estimators | [20, 50, 100, 300, 400, 500, 1000] | 20 |
|  | Learning rate | [0.001, 0.01, 0.05, 0.1, 0.5] | 0.01 |
| XGBoost | Number of estimators | [20, 50, 100, 300, 400, 500, 1000] | 100 |
|  | Max depth | [1, 3, 5, 6, 7, 10] | 1 |
|  | Eta | [0.01, 0.03, 0.05, 0.1, 0.2] | 0.1 |
|  | Min child weight | [0.1, 0.3, 0.5] | 0.3 |
|  | Max leaf nodes | [4, 6, 9, 10] | 9 |
|  | Subsample | [0.1, 0.5, 0.8, 1] | 0.5 |
|  | Gamma | [0.01, 0.05, 0.1, 0.2, 0.5, 0.8] | 0.1 |
|  | Alpha | [0.001, 0.01, 0.1, 0.5] | 0.5 |
|  | Max delta step | [0, 1, 2, 5] | 2 |
|  | Colsample by tree | [0.5, 0.6, 0.8] | 0.5 |
|  | Colsample by level | [0.4, 0.6, 0.8] | 0.6 |
|  | Colsample by node | [0.2, 0.3, 0.5, 0.8] | 0.3 |
|  | Lambda | [0.01, 0.05, 0.1, 0.3, 0.5] | 0.05 |

Table 3. MI mortality prediction performance of ensemble tree-based ML models.

| Algorithm | C-Statistic (AUC) | Accuracy | Sensitivity | Specificity | Precision |
|---|---|---|---|---|---|
| RF | 0.83 ± 0.10 *** | 77 ± 9 % *** | 88 ± 11 % *** | 60 ± 20 % * | 79 ± 9 % ** |
| AdaBoost | 0.82 ± 0.11 *** | 78 ± 10 % *** | 85 ± 13 % *** | 67 ± 20 % *** | 82 ± 9 % *** |
| XGBoost | 0.80±0.11 * | 76 ± 9 % *** | 82 ± 13 % * | 67 ± 21 % *** | 81 ± 10 % *** |
| LR | 0.77±0.13 | 71 ± 10 % | 81 ± 12 % | 56 ± 23 % | 76 ± 10 % |

P-values are obtained by a paired t-test in comparison to the baseline LR mode. *, **, and *** imply p-values under 0.05, 0.01, and 0.001, respectively.

Table 4. Effect of bPEP and bET on the prediction performance of different ML models.

| Experiment | Algorithm | C-statistic (AUC) | Accuracy | Sensitivity | Specificity | Precision |
|---|---|---|---|---|---|---|
| Experiment I (With all features) | RF | 0.83 *** | 77% *** | 88% ** | 60% *** | 79% *** |
|  | AdaBoost | 0.82 *** | 78% *** | 85% *** | 67% | 82% |
|  | XGBoost | 0.80 * | 76% | 82% | 67% | 81% |
| Experiment II (bET & bPEP excluded) | RF | 0.80 | 73% | 86% | 52% | 75% |
|  | AdaBoost | 0.79 | 78% | 85% | 67% | 81% |
|  | XGBoost | 0.78 | 77% | 83% | 67% | 81% |

*, **, and *** imply p-values under 0.05, 0.01, and 0.001, respectively

Supplementary Table 1. MI mortality prediction performance on train dataset of ensemble tree-based ML models for Experiment I (including all features).

| Algorithm | C-Statistic (AUC) | Accuracy | Sensitivity | Specificity | Precision |
|---|---|---|---|---|---|
| RF | 0.83 ± 0.11 | 88 ± 4 % | 93 ± 2 % | 79 ± 8 % | 88 ± 4 % |
| AdaBoost | 0.82 ± 0.10 | 84 ± 2 % | 90 ± 2 % | 74 ± 5 % | 85 ± 3 % |
| XGBoost | 0.81 ± 0.12 | 88 ± 6 % | 93 ± 2 % | 78 ± 14 % | 88 ± 6 % |
| LR | 0.77 ± 0.13 | 77 ± 2 % | 85 ± 1 % | 64 ± 4 % | 80 ± 2 % |

Supplementary Table 2. MI mortality prediction performance on train dataset of ensemble tree-based ML models for Experiment II (Excluding bET & bPEP).

| Algorithm | C-Statistic (AUC) | Accuracy | Sensitivity | Specificity | Precision |
|---|---|---|---|---|---|
| RF | 0.81 ± 0.11 | 83 ± 5 % | 91 ± 2 % | 69 ± 14 % | 83 ± 5 % |
| AdaBoost | 0.79 ± 0.13 | 81 ± 3 % | 86 ± 4 % | 73 ± 5 % | 84 ± 3 % |
| XGBoost | 0.77 ± 0.12 | 80 ± 8 % | 91 ± 4 % | 60 ± 26 % | 79 ± 9 % |
| LR | 0.77 ± 0.13 | 79 ± 2 % | 86 ± 2 % | 68 ± 4 % | 82 ± 2 % |